\documentclass[letterpaper, 10 pt, conference]{ieeeconf}
\usepackage[dvipsnames]{xcolor} 
\usepackage{algorithm}
\usepackage{algpseudocode}
\usepackage{algorithmicx}

\usepackage{array}
\usepackage{textcomp}
\usepackage{stfloats}
\usepackage{changes}
\usepackage{subcaption}
\usepackage{url}
\usepackage{verbatim}
\usepackage{graphicx}
\usepackage{cite}
\usepackage{hyperref}

\usepackage{amsmath, amsfonts, amssymb, amsthm}
\usepackage{mathtools}
\usepackage{ulem}
\usepackage{booktabs}

\usepackage{siunitx}    

\usepackage[T1]{fontenc} 

\IEEEoverridecommandlockouts
\hypersetup{hidelinks}
\usepackage{multirow}
\usepackage{multicol}
\usepackage[table]{xcolor}
\begin{document}
\title{\LARGE \bf Vision-Guided MPPI for Agile Drone Racing: Navigating Arbitrary Gate Poses via Neural Signed Distance Fields}

\author{Fangguo Zhao$^{1}$, Hanbing Zhang$^{1}$, Zhouheng Li$^{1}$, Xin Guan$^{1}$, and Shuo Li$^{1}$
\thanks{$^{1}$Authors are with the College of Control Science and Engineering, Zhejiang University, Hangzhou 310027, China {\tt\small shuo.li@zju.edu.cn}}%
}
\maketitle

\begin{abstract}
Autonomous drone racing requires the tight coupling of perception, planning, and control under extreme agility. However, recent approaches typically rely on precomputed spatial reference trajectories or explicit 6-DoF gate pose estimation, rendering them brittle to spatial perturbations, unmodeled track changes, and sensor noise. Conversely, end-to-end learning policies frequently overfit to specific track layouts and struggle with zero-shot generalization. To address these fundamental limitations, we propose a fully onboard, vision guided optimal control framework that enables reference-free agile flight through arbitrarily placed and oriented gates. Central to our approach is Gate-SDF, a novel, implicitly learned neural signed distance field. Gate-SDF directly processes raw, noisy depth images to predict a continuous spatial field that provides both collision repulsion and active geometric guidance toward the valid traversal area. We seamlessly integrate this representation into a sampling-based Model Predictive Path Integral (MPPI) controller. By fully exploiting GPU parallelism, the framework evaluates these continuous spatial constraints across thousands of simulated trajectory rollouts simultaneously in real time. Furthermore, our formulation inherently maintains spatial consistency, ensuring robust navigation even under severe visual occlusion during aggressive maneuvers. Extensive simulations and real-world experiments demonstrate that the proposed system achieves high-speed agile flight and successfully navigates unseen tracks subject to severe unmodeled gate displacements and orientation perturbations. Videos are available at \url{https://zhaofangguo.github.io/vision_guided_mppi/}
\end{abstract}

\section{Introduction}

Autonomous quadrotors have recently achieved record-breaking agile flight performance, enabling applications previously deemed unattainable \cite{10530312}. Autonomous drone racing, in particular, has emerged as a compelling benchmark that stresses perception, planning, and control under aggressive, near time-optimal maneuvers. The primary objective is to navigate a sequence of gates in a predefined order. While numerous studies have tackled this task, the majority simplify gates into sequences of spatial waypoints, which stripping away critical geometric constraints or overfit their policies to specific, static track layouts. The ability to fly like a human pilot, freely reacting to unstructured tracks and arbitrarily oriented gates without prior map knowledge, remains a largely unexplored frontier. In this paper, we propose a vision guided optimal control framework capable of racing through arbitrary gate configurations without relying on predefined reference trajectories or precise global gate poses.

\begin{figure}[t]
    \centering
    \includegraphics[width=0.49\textwidth]{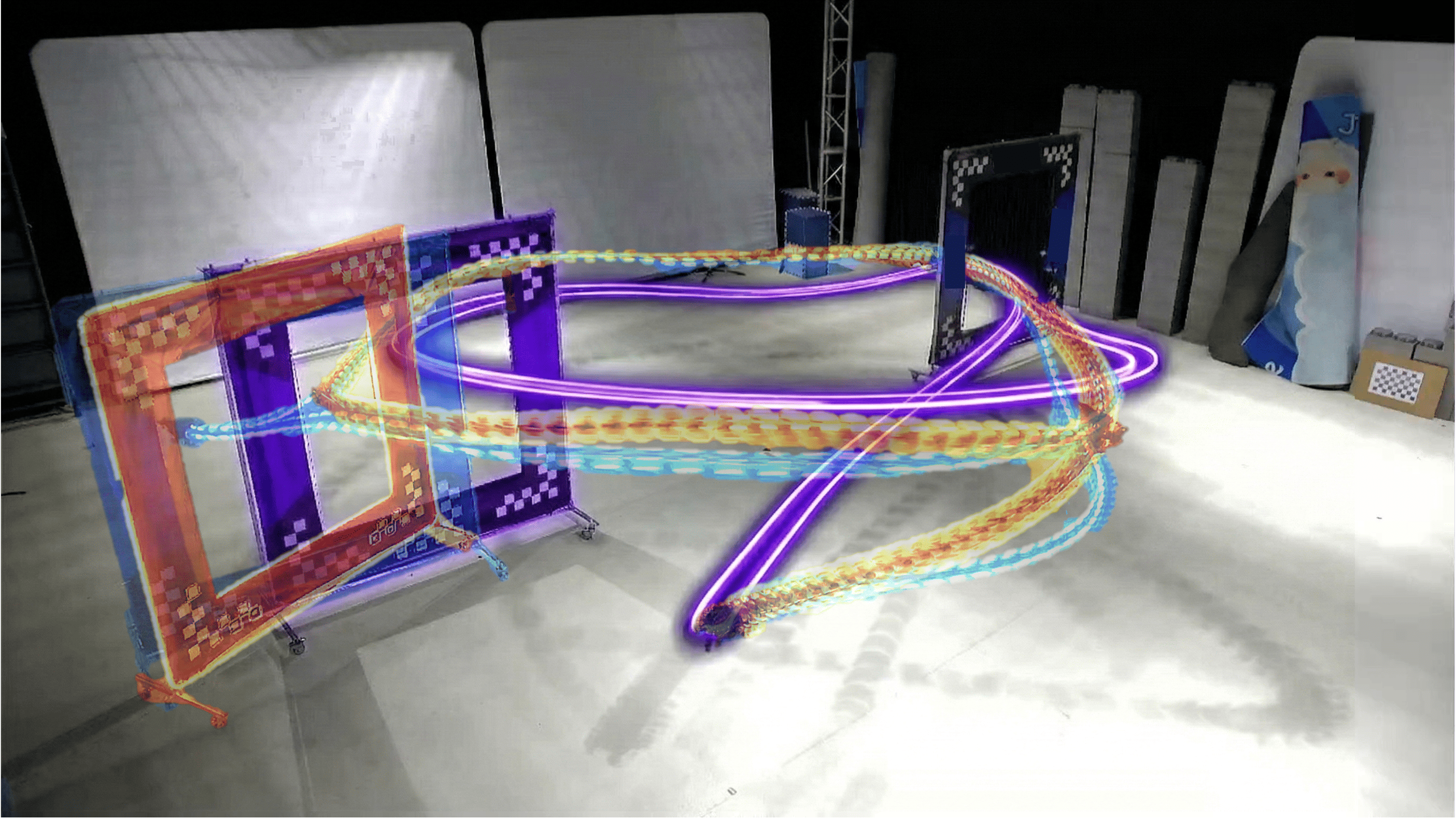}
    \caption{Real-world experiments with gates arranged at varying positions and orientations. All computations are performed online by the onboard computer using depth images.}
    \label{fig:realtraj}
    \vspace{-2em}
\end{figure}

Traditional model-based optimal control for drone racing often treats sequential gate traversal as a pure waypoint-tracking problem, heavily relying on explicit reference trajectories. Early methods utilized polynomials to plan smooth paths through predefined waypoints \cite{mellinger2011minimum, 7762111,10342456}. Subsequent works advanced this by employing full-state nonlinear optimization to generate time-optimal reference trajectories \cite{foehn2021time,10341844}, or by using simplified point-mass models to efficiently compute geometric references \cite{romero2022model}. However, tracking these precomputed trajectories inherently serves as a proxy task rather than directly optimizing for racing performance \cite{song2023reaching}. To bypass the reliance on explicit references, recent learning-based methods have been introduced \cite{11246459}; yet, they struggle to generalize across the entire agile flight envelope and still require auxiliary polynomial planning to ensure stability. Recent works adapt control policies to new tracks by modeling the environment itself, but this requires training an additional network \cite{wang2025environmentpolicylearningrace}. More recently, a sampling-based control approach demonstrated fully reference-free multi-waypoint flight \cite{zhao2025rethinkingreferencetrajectoriesagile}, providing a principled path toward reference-free racing within an optimal control framework.

To improve upon pure waypoint tracking, subsequent research incorporated the physical dimensions of the gates directly into the planning formulation. These approaches enforce spatial constraints specifically at the gate locations \cite{10610148,10342456,9543598} or along the whole track\cite{krinner2024mpccmodelpredictivecontouring}, or utilize analytical spatial guidance \cite{11127454}. While effective in structured settings, these methods fundamentally assume that the environment\cite{9543598} or the optimal passing state\cite{10342456} is fully known and static\cite{krinner2024mpccmodelpredictivecontouring}. They lack the robustness to handle spatial perturbations or changing gate poses, and generating trajectories for a novel track layout remains a computationally expensive, offline process.

A central challenge in unstructured drone racing is perceiving and reacting to gates online using strictly onboard sensing. Early approaches explicitly estimated the 6-DoF gate pose using traditional \cite{7989679,li2020autonomous,li2020visual} or learning-based \cite{kaufmann2023champion,Bosello_2026,azhari2025driftcorrectedmonocularvioperceptionaware,bahnam2026monoracewinningchampionleveldrone} corner detection coupled with Perspective-n-Point (PnP) geometry. However, these methods require exact prior knowledge of gate dimensions and the simultaneous detection of multiple corners. Consequently, they are highly sensitive to motion blur and occlusions, leading to fragile and noisy localization during aggressive maneuvers.

To bypass the fragility of explicit geometric solvers, implicit and end-to-end learning methods have been explored. Initial efforts regressed gate poses directly from RGB images \cite{kaufmann2019beauty}, while recent reinforcement learning (RL) policies map raw pixels directly to motor commands \cite{geles2024demonstratingagileflightpixels,xing2024bootstrapping}. Despite their agility, these vision-based RL policies frequently overfit to specific tracks and struggle to generalize to unseen layouts or arbitrary gate orientations. Although using depth images improves RL generalization \cite{11297752}, it typically trades time-optimal speed for conservative robustness.

To overcome these limitations, we propose a fully onboard, vision guided racing framework driven by depth observations. By integrating a learned geometric representation with a Model Predictive Path Integral (MPPI) controller, our approach eliminates the need for explicit reference trajectories and gate feature detection, relying solely on sparse target waypoints and real-time depth images.

\begin{figure*}[t]
    \centering
    \includegraphics[width=0.99\textwidth]{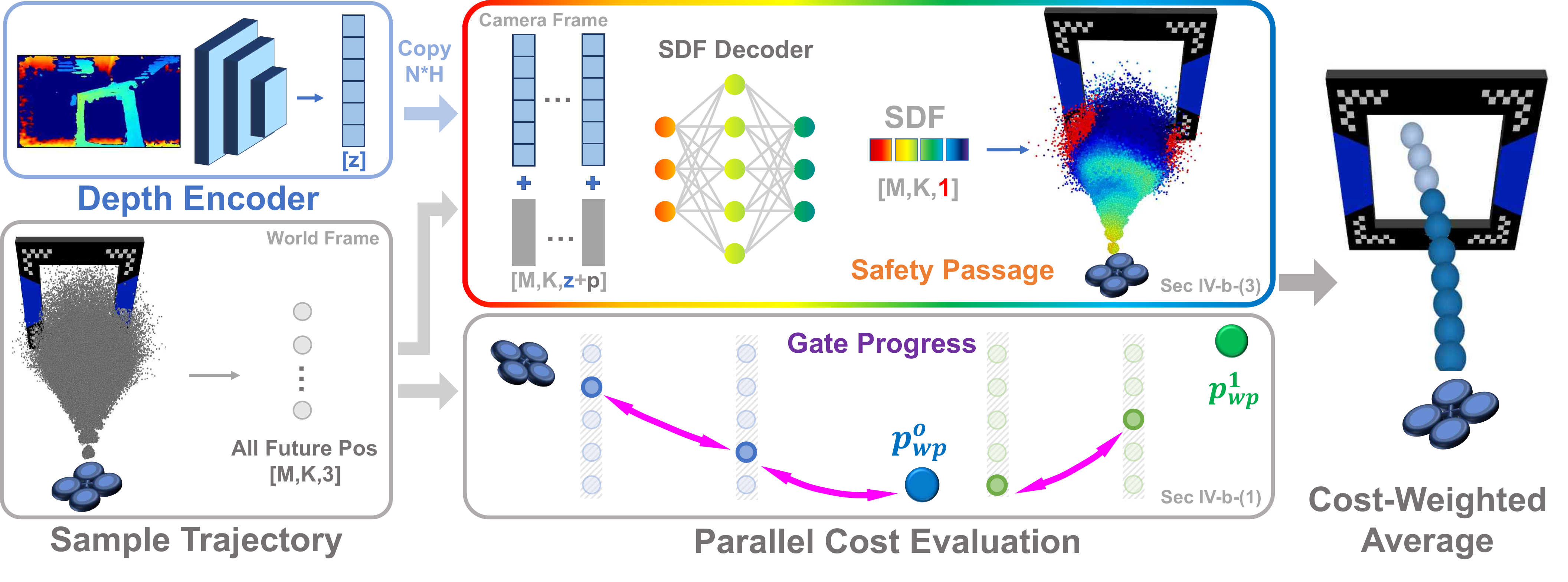}
    \caption{Overview of the proposed framework. At each control step, a depth encoder extracts a latent vector $\mathbf{z}$, which is duplicated $M \times K$ times ($M$ rollouts, $K$ horizon) and concatenated with each MPPI-sampled state $\mathbf{p}$. The SDF decoder evaluates these points to formulate vision guided safety constraints. Combined with a gate progress objective, the optimal control sequence is derived via cost-weighted averaging.}
    \label{fig:main}
    \vspace{-1em}
\end{figure*}

The main contributions of this paper are as follows:
\begin{enumerate}
    \item A novel, vision guided framework capable of navigating tracks with arbitrary gate positions and orientations using onboard vision, completely eliminating the need for predefined reference trajectories.
    \item A tightly coupled Neural Signed Distance Field (SDF) and MPPI control architecture. This integration efficiently embeds complex spatial constraints into optimal control by fully leveraging parallel GPU computation for real-time trajectory sampling.
    \item Extensive validation in both simulations and real-world experiments, demonstrating that the proposed SDF-enhanced MPPI framework achieves robust and agile flight across diverse and unseen track configurations.
\end{enumerate}

\section{Related Work}
A principled method to incorporate perception into model-based control is to represent the environment using Signed Distance Fields (SDFs)\cite{8954065}, which map spatial coordinates to the distance of the nearest obstacle. While effective for general collision avoidance, traditional SDFs only encode the shortest distance to generic surfaces. Consequently, for highly structured and porous objects like racing gates, they fail to differentiate between the traversable interior opening and the non-traversable exterior frame. Recent works have attempted to learn implicit SDFs via neural networks and integrate them as collision constraints within Model Predictive Control (MPC) pipelines \cite{jacquet2025neural}. However, standard gradient-based MPC solvers require specifically tailored architectures to ensure strict network differentiability, thereby imposing significant limitations and rigid constraints on the overall network design.

To overcome the limitations of gradient-based optimizers, Model Predictive Path Integral (MPPI) control offers a sampling-based, gradient-free alternative. Instead of computing analytical gradients, MPPI optimizes control sequences by simulating thousands of state rollouts and evaluating their trajectory costs in parallel, allowing for the straightforward integration of highly non-convex constraints. Existing MPPI approaches typically achieve obstacle avoidance by penalizing collisions using generic environmental representations, such as OctoMaps \cite{minavrik2024model}, 3D occupancy grids \cite{zhai2025pa}, spherical grids \cite{chen2025aero}, or direct depth-image projections \cite{pochobradský2025geometricmodelpredictivepath}. While these variants successfully avoid general obstacles, they lack the geometric understanding required to actively navigate \textit{through} structured constraints like arbitrarily oriented racing gates. In contrast, our work represents gates with a learned Neural SDF and combines it with the highly parallel rollout capabilities of MPPI, enabling active, geometry-aware navigation through narrow openings.

\section{Gate-Shaped Signed Distance Field}
In this section, we introduce the construction of a gate-centric neural signed distance field, termed Gate-SDF. We also detail the data generation pipeline and the network training procedure.

\subsection{Gate-specific signed distance field (Gate-SDF)}
To overcome the limitations of traditional Signed Distance Fields (SDFs), which cannot distinguish between the safe opening and the solid frame of racing gates, we design an analytical spatial representation termed Gate-SDF. Let $\mathbf{p} = [x, y, z]^\top$ represent a 3D query point in the local gate coordinate frame, where the gate plane lies at $x=0$. We define a radial distance metric $r(\mathbf{p})$ tailored to the specific gate geometry, utilizing the $\ell_{\infty}$ norm on the transverse plane for standard square gates:
\begin{equation}
    r(\mathbf{p}) = \max(|y|, |z|).
\end{equation}

We construct the safety area as an hourglass-shaped frustum that creates a funneling effect toward the gate center, as illustrated in Fig.~\ref{fig:gate_sdf-a}. Let $c$ denote the half-width of the inner gate opening, and $\alpha$ represent a predefined cone expansion angle. The guidance SDF, $s_{\text{guide}}(\mathbf{p})$, is defined as:
\begin{equation}
    s_{\text{guide}}(\mathbf{p}) = c + |x|\tan(\alpha) - r(\mathbf{p}).
\end{equation}
Under this formulation, $s_{\text{guide}}(\mathbf{p}) > 0$ indicates a state within the safe area, while $s_{\text{guide}}(\mathbf{p}) < 0$ implies a deviation from the safe approach area. This linear expansion along $|x|$ ensures that the allowable spatial margin increases as the distance from the gate plane increases, providing a forgiving gradient that actively guides the drone into the narrow traversal window.

\subsection{Learning the Safety Region without External Gate Pose}
\begin{figure}[h]
    \centering
    \begin{subfigure}{0.49\linewidth}
        \centering
        \includegraphics[width=\linewidth, trim=4cm 2cm 4cm 2cm, clip]{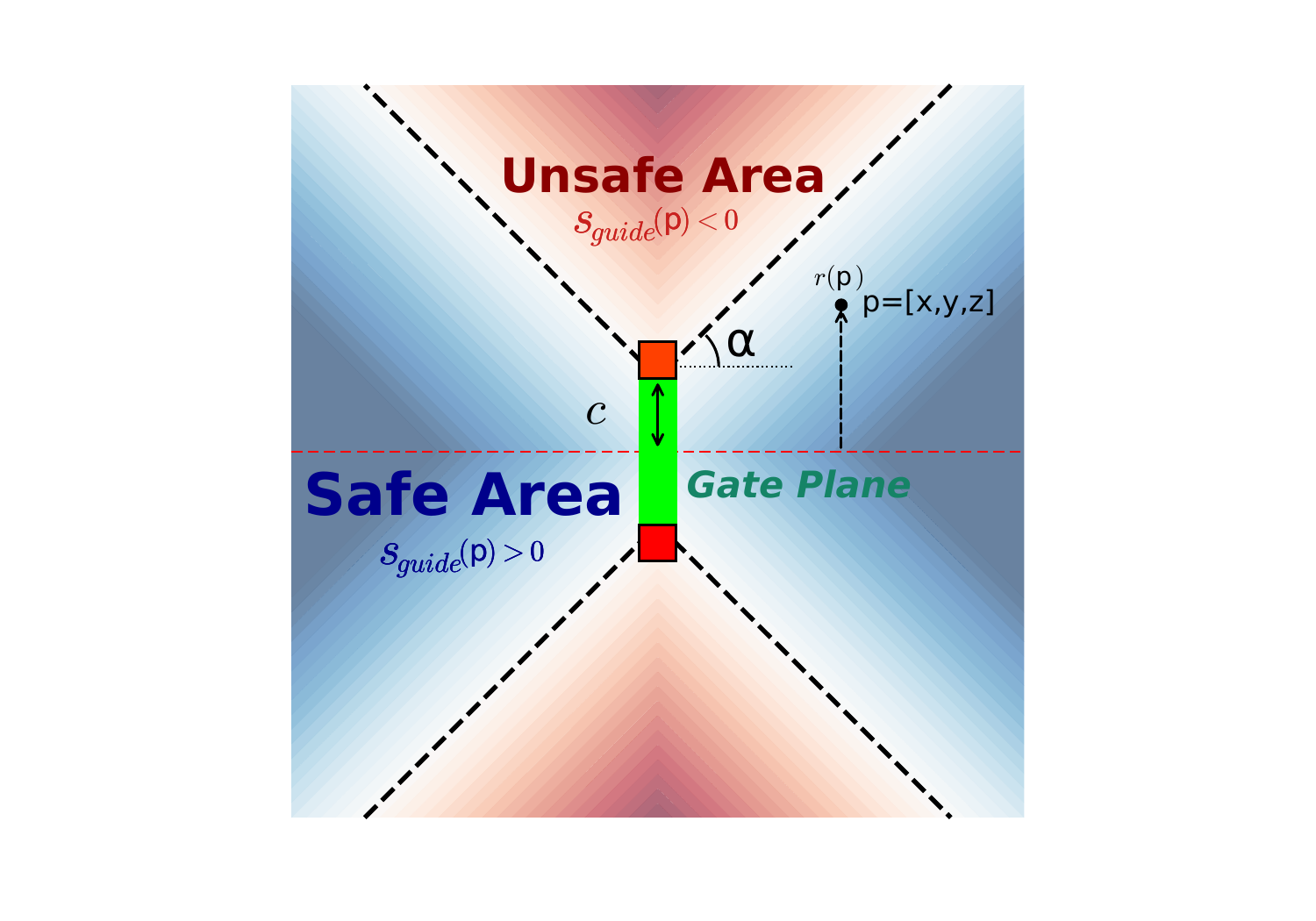}
        \caption{Analytical construction of the Gate-SDF.}
        \label{fig:gate_sdf-a}

    \end{subfigure}\hfill
    \begin{subfigure}{0.49\linewidth}
        \centering
        \includegraphics[width=\linewidth]{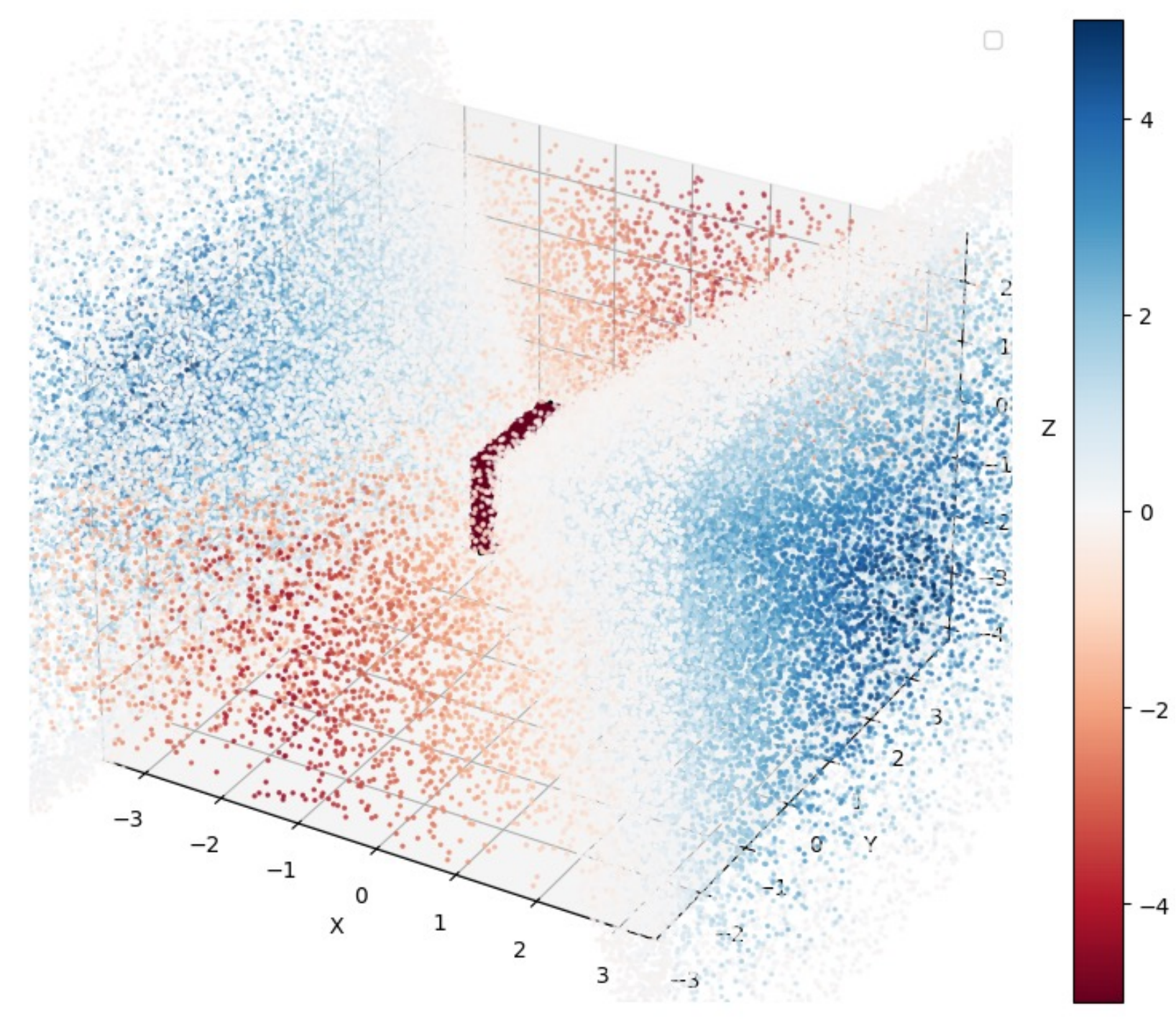}
        \caption{Hybrid spatial sampling strategy.}
        \label{fig:gate_sdf-b}
    \end{subfigure}
    
    \vspace{1ex}
    \begin{subfigure}{\linewidth}
        \centering
        \includegraphics[width=\linewidth]{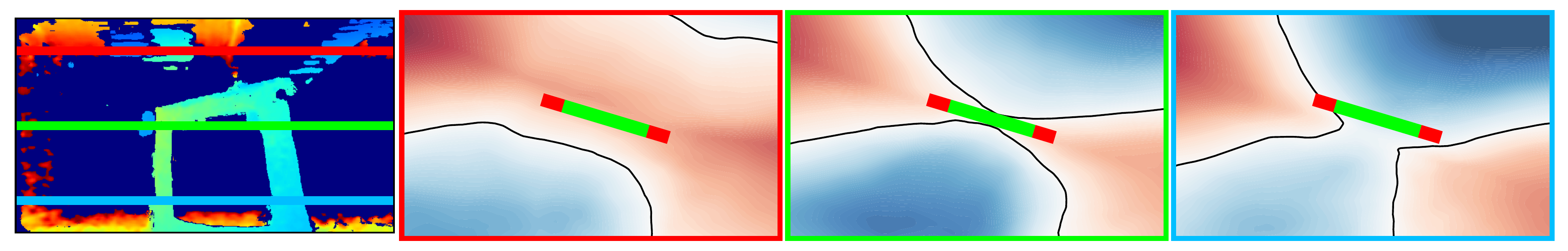}
        \caption{Predicted Gate-SDF visualization. The leftmost panel is the onboard depth image. Three horizontal slices (colored lines) are sampled, and their top-down SDF maps show that the model correctly captures the gate's traversable region.}
        \label{fig:gate_sdf-c}
    \end{subfigure}
    
    \caption{Dataset Generation Pipeline.}
\end{figure}
A key challenge in autonomous drone racing is accurately identifying the traversable safety region during flight without relying on external motion capture systems or \textit{a priori} maps. While the Gate-SDF effectively represents this safe area, traditional pipelines require explicitly estimating the 6-DoF gate pose during flight to transform the predefined SDF into the current frame. However, directly estimating the full 6-DoF pose from onboard sensing is highly unreliable under motion blur, severe occlusions, and aggressive maneuvers. To circumvent this limitation, we design a lightweight Gate-SDF network that implicitly learns the geometric safety region directly from onboard depth observations. Specifically, given a first-person-view (FPV) depth image and an arbitrary 3D spatial query point defined in the camera coordinate frame, our network directly predicts the corresponding SDF value at that location. In the remainder of this section, we detail the data generation pipeline and the two-stage training procedure for the proposed network.

To generate robust training supervision, we utilize the ground-truth Gate-SDF as the target. To ensure the network learns the precise boundaries, we employ a hybrid spatial sampling strategy during data generation, as shown in Fig.~\ref{fig:gate_sdf-b}:
\textbf{Near-Surface Samples:} Points sampled densely around the theoretical boundary ($s_{\text{guide}} \approx 0$) to refine zero-crossing accuracy. \textbf{Interior Samples:} Points strictly within the safe area to learn the continuous distance values and guidance gradients. \textbf{Collision-Prone Samples:} Points sampled within and near the physical gate frame to enforce strict obstacle avoidance. \textbf{Global Uniform Samples:} Points distributed throughout the camera frustum to provide spatial background context.

To cover diverse viewpoints, we randomly sample the quadrotor pose in the space. For each sampled pose, we render and record the corresponding depth image using Genesis\cite{Genesis}. We then sample 3D query points within the camera frame, evaluate the analytic Gate-SDF at these points, and downsample to $8192$ point--SDF pairs per image.

\begin{figure}[h]
    \centering
    \includegraphics[width=0.49\textwidth]{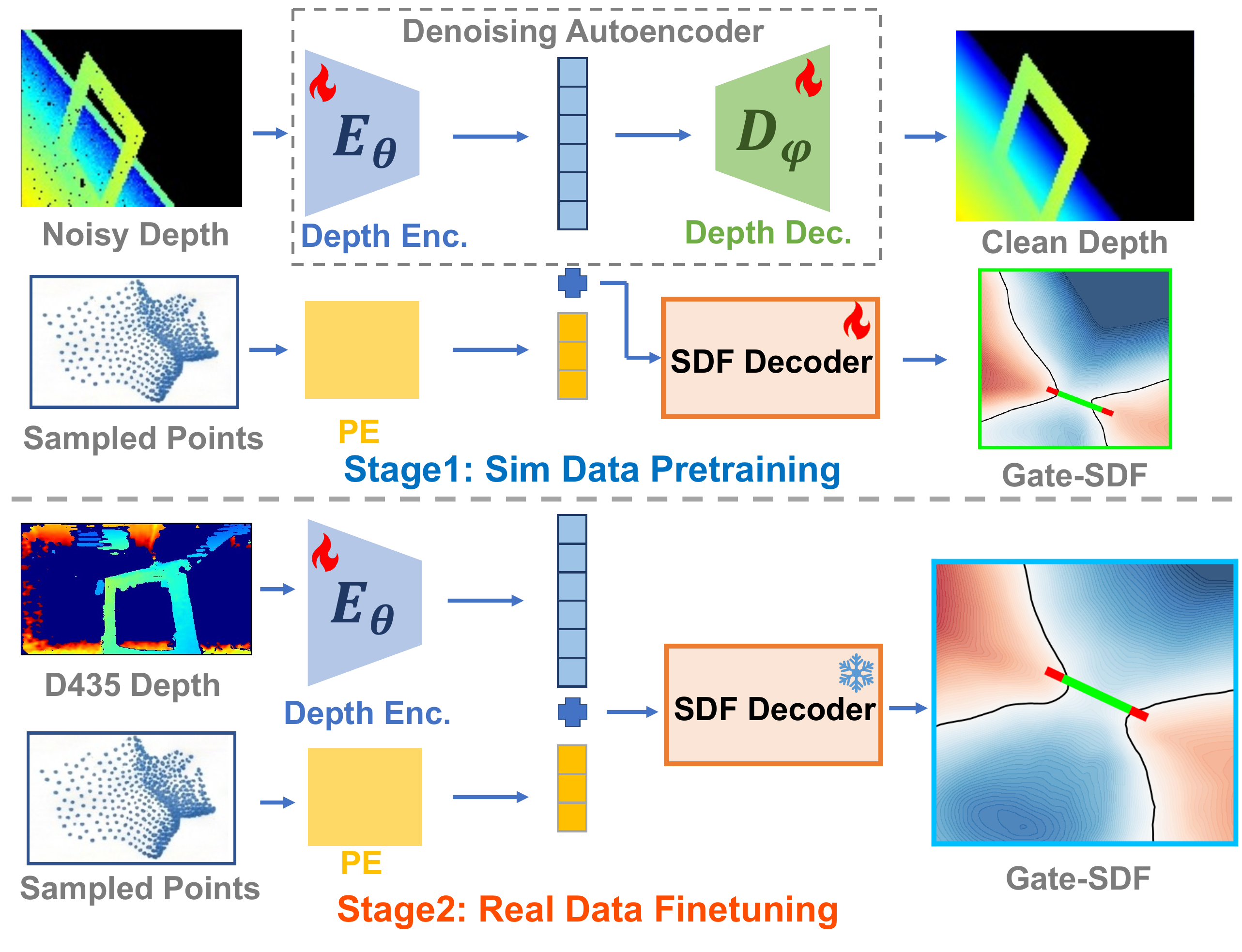}
    \caption{Overview of the two-stage training pipeline.
     In simulation, a denoising autoencoder is trained to robustly extract gate features from noisy images, alongside a SDF decoder. In the real-world domain, the image encoder is fine-tuned to adapt to specific sensor characteristics.}
    \label{fig:network}
    \vspace{-1em}
\end{figure}

To bridge the sim-to-real gap caused by physical sensor artifacts in real-world depth images, we propose a two-stage training paradigm for our lightweight depth-conditioned SDF network. As shown in Fig.\ref{fig:network}, the architecture consists of a depth encoder $E_{\theta}$, an auxiliary depth decoder $D_{\psi}$, and an MLP SDF decoder $g_{\phi}$.

In the first stage (Simulation Pre-training), we train the full network on a synthetic dataset with $50000$ image-SDF pairs. To improve robustness, we apply synthesized hardware noise to the clean simulated depth $\mathcal{I}_{\text{clean}}$ to generate a noisy input $\mathcal{I}_{\text{noisy}}$. The encoder maps this input to a latent code $\mathbf{z}=E_{\theta}(\mathcal{I}_{\text{noisy}})$. The network jointly performs denoising depth reconstruction $\hat{\mathcal{I}}_{\text{clean}} = D_{\psi}(\mathbf{z})$ and signed distance prediction $\hat{s} = g_{\phi}(\mathbf{z}, \text{PE}(\mathbf{p}))$ for a query point $\mathbf{p}\in\mathbb{R}^3$ in the camera frame, where $\text{PE}(\cdot)$ denotes a positional encoding applied to the spatial coordinates before concatenation. The network is trained using an $\mathcal{L}_1$ loss formulation:
\begin{equation}
    \mathcal{L}_{\text{stage1}} = \lambda_{\text{recon}} \|\hat{\mathcal{I}}_{\text{clean}} - \mathcal{I}_{\text{clean}}\|_1 + \lambda_{\text{sdf}} \|\hat{s} - s_{\text{guide}}(\mathbf{p})\|_1.
\end{equation}
where $\lambda_{\text{recon}}$ and $\lambda_{\text{sdf}}$ are weighting parameters. This stage functions as a Denoising Autoencoder (DAE), enabling the encoder to extract noise-invariant representations while the SDF decoder masters the gate geometry across diverse configurations.

In the second stage (Real-world Fine-tuning), we adapt the network to real sensor data. To construct this real-world dataset, we recorded the quadrotor's pose and the gate's position within a motion capture environment, collecting approximately one minute of flight data with depth images captured at $60\,\text{Hz}$, yielding $3000$ image-SDF pairs. Since perfectly clean ground-truth depth is unavailable in the real world, the reconstruction task is dropped. We freeze the weights of the well-trained SDF decoder $g_{\phi}$ and depth decoder $D_{\psi}$, and exclusively fine-tune the encoder $E_{\theta}$ on real-world noisy depth images. The fine-tuning is guided strictly by the geometric supervision:
\begin{equation}
    \mathcal{L}_{\text{stage2}} = \|\hat{s} - s_{\text{guide}}(\mathbf{p})\|_1.
\end{equation}
By freezing the SDF decoder as a "teacher", the "student" encoder is forced to map real noisy inputs into the established clean latent space. At runtime, $D_{\psi}$ is discarded, and the value of Gate-SDF for each position in the field of view (FOV) can then be efficiently evaluated by transforming the position into the image frame and forwarding it alongside the latent code through $g_{\phi}$, as shown in Fig.\ref{fig:gate_sdf-c}.

\section{Vision Guided Racing Controller via MPPI}
The Gate-SDF introduced above provides a continuous and dense cost map conditioned on a single depth image. Given a 3D position expressed in the image frame, the Gate-SDF network outputs its signed distance to a predefined safety margin, which serves as a metric to evaluate whether a state is safe or unsafe. We integrate the network into sampling-based MPC, specifically MPPI, where costs are evaluated across a batch of sampled trajectories. This pairing naturally leverages hardware parallelism: batched network inference can score all rollout states simultaneously on the GPU. In this section, we first present the core principles of MPPI and then detail the task-driven cost function used in our MPPI formulation for drone racing.

\subsection{Model Predictive Path Integral Control Framework}
\label{sec:mppi}
Model Predictive Path Integral (MPPI) control optimizes control sequences via sampling-based evaluation \cite{williams2018information}. At each control step, $M$ candidate control sequences $\mathbf{U}^{m} = \mathbf{U}^* + \delta\mathbf{U}^{m}$ are generated by injecting Gaussian noise $\delta\mathbf{u}_k^{m} \sim \mathcal{N}(0, \Sigma)$ into the previous optimal sequence $\mathbf{U}^*$. Each candidate is rolled out using the system dynamics to obtain a state trajectory, and its total cost $\mathcal{J}(\mathbf{U}^{m})$ is evaluated. The optimal sequence is then updated as a cost-weighted average:
\begin{equation}
    \mathbf{U}^* = \sum_{m=1}^{M} w_m \mathbf{U}^{m}, \quad w_m = \frac{\exp\left(-\frac{1}{\lambda} \mathcal{J}(\mathbf{U}^{m})\right)}{\sum_{i=1}^{M} \exp\left(-\frac{1}{\lambda} \mathcal{J}(\mathbf{U}^{i})\right)},
    \label{eq:mppi_update}
\end{equation}
where $\lambda > 0$ is the temperature parameter. Only the first action of $\mathbf{U}^*$ is applied, and the sequence initializes the next control cycle. 

For the forward simulation of the rollouts, we employ a standard quadrotor dynamics model $\dot{\mathbf{x}} = f(\mathbf{x}, \mathbf{u})$. The system state $\mathbf{x} = [\mathbf{p}, \mathbf{v}, \mathbf{q}, \boldsymbol{\omega}]^T$ comprises the position, linear velocity, orientation (quaternion), and angular velocity. Following \cite{romero2022model}, the control input $\mathbf{u} = [\dot{T}_1, \dot{T}_2, \dot{T}_3, \dot{T}_4]^T$ is defined as the derivative of the individual rotor thrusts to account for motor dynamics and ensure smooth control signals.

\subsection{Cost for Drone Racing}
\subsubsection{Gate Progress Cost}
To push the drone to navigate aggressively through a sequence of waypoints while safely passing each gate, we also adopt a task-centric objective that directly relates to the racing goal, i.e., maximizing flight progress toward the current target waypoint. The gate progress cost directly optimizes progress towards sequential waypoints without an explicit reference path:
\begin{equation}
\begin{aligned}
    \mathcal{J}_{gate} = \sum_{k=0}^{K-1} \left(\|\mathbf{p}_{k+1} - \mathbf{p}_{\text{wp}}\|_2^2 - \|\mathbf{p}_{k} - \mathbf{p}_{\text{wp}}\|_2^2\right),
\end{aligned}
\label{eq:gate_progress}
\end{equation}
where $\mathbf{p}_{\text{wp}}$ denotes the position of the current target waypoint.

\subsubsection{Perception Alignment Cost}
To ensure the target gate remains within the field of view of the forward-facing depth camera during aggressive maneuvers, we introduce a perception-alignment cost. This cost actively aligns the quadrotor's yaw angle $\psi_k$ with the line-of-sight angle to the gate center $\psi_{\text{gate},k}$ at each time step $k$:
\begin{equation}
    \mathcal{J}_{\text{vis}} = \sum_{k=0}^{K-1} \left( 1 - \cos(\psi_k - \psi_{\text{gate},k}) \right).
    \label{eq:visibility_cost}
\end{equation}
Minimizing this cost continuously directs the camera toward the upcoming gate, thereby maximizing observability and stabilizing onboard perception.

\subsubsection{Gate-SDF Guided Safety Cost}
The learned SDF is incorporated into MPPI as an additional vision guided cost that penalizes low-clearance or unsafe rollout states. At the beginning of each control cycle, we encode the current depth image $\mathcal{I}_{\text{depth}}$ once to obtain a latent vector $\mathbf{z}$. During each rollout, we query the decoder for every predicted position $\mathbf{p}_k$ and obtain the SDF value $\hat{s}_k = g_{\phi}(\mathbf{z},\mathbf{p}_k)$ in a single forward evaluation.

For a predicted position $\mathbf{p}_k$ with SDF value $\hat{s}_k$, we define the SDF stage cost as
\begin{equation}
    C_{\text{sdf}}(\mathbf{p}_k,\mathbf{z}) = \max\big(0,\, d_{\text{safe}}-\hat{s}_k\big),
\end{equation}
where $d_{\text{safe}}\ge 0$ is the desired clearance. The total SDF cost over the horizon is then
\begin{equation}
\begin{aligned}
    \mathcal{J}_{\text{sdf}} = \sum_{k=0}^{K-1} C_{\text{sdf}}(\mathbf{p}_k,\mathbf{z}).
\end{aligned}
\end{equation}

During aggressive maneuvers, the target gate frequently leaves the camera's field of view (FOV). Our Gate-SDF addresses this through inherent \textit{spatial consistency}, endowing the system with \textit{object permanence}. Since the inferred 3D gate geometry remains structurally invariant within its local frame, we continuously cache the latest valid latent vector and its corresponding world-to-camera transformation. Under visual occlusion, the system evaluates SDF values for MPPI rollouts by transforming world-frame query points into the cached camera frame and forwarding them through the decoder using the stored latent code. This persistent spatial memory ensures robust navigation even without continuous visual feedback.

Finally, the vision guided objective used by MPPI is the sum of the progress-related cost and the SDF safety cost:
\begin{equation}
    \mathcal{J} = Q_{\text{gate}} \mathcal{J}_{\text{gate}} + Q_{\text{vis}} \mathcal{J}_{\text{vis}} + Q_{\text{sdf}} \mathcal{J}_{\text{sdf}}.
\end{equation}
where, $Q_{\text{gate}}$, $Q_{\text{vis}}$, and $Q_{\text{sdf}}$ are scalar weights that balance progress, visibility alignment, and SDF safety, respectively.

\section{Experiments}
We validate the proposed vision guided, reference-free controller through extensive simulations and real-world flight experiments. In simulation, we construct race tracks with varying difficulty levels and systematically evaluate different combinations of gate positions and orientations. Throughout all trials, the actual gate pose remains unknown to the system: the controller is provided only with nominal waypoint coordinates, while all gate interactions, including approach, alignment, and collision avoidance rely entirely on onboard depth observations.

To ensure a consistent and rigorous benchmark across both simulation and hardware, we utilize gates with identical dimensions in all scenarios. Specifically, each gate features an inner diameter of $1.0 \text{ m}$ and an outer diameter of $1.5 \text{ m}$. This narrow opening imposes a stringent spatial constraint for high-speed flight, thoroughly stressing the robustness and precision of our method.

\subsection{Simulation Experiments}
\subsubsection{Implementation Details}
\begin{table}[h]
    \centering
     \caption{The parameters used in simulation}
    \begin{tabular}{cc|cc}
     \toprule
    \textbf{Parameters} & \textbf{Value/Range} & \textbf{Parameters} & \textbf{Value/Range} \\
    \midrule
        $M$ & $8192$ & $\lambda$ & $[0.01,0.1]$ \\  
        $K$ & $20$ & $\Delta t$ & $0.03$ \\
         $\dot{T_s}$ & $[-10,10]$ & $d_{\text{safe}}$ & $1.0$\\ 
        $\omega[rad/s]$ & $[-10,10]$ & $T_{s}[N]$ & $[0,10]$\\ 
         \bottomrule
    \end{tabular}
     \label{tab:mppi_param}
    \vspace{-1.5em}
\end{table}
Our MPPI controller is implemented in \texttt{JAX} to exploit highly optimized, GPU-parallel computation, drawing inspiration from the architecture in \cite{yi2024covo}. All simulation experiments are executed on an NVIDIA RTX 3090 GPU. The solving time for a single control step averages approximately $3 \text{ ms}$, comfortably satisfying real-time high-frequency control requirements. We utilize \textit{Genesis} as the physics and rendering engine, which streams onboard depth observations to the controller.

The Gate-SDF network is trained offline using \texttt{PyTorch} with latent vector $\mathbf{z} \in \mathbb{R}^{128}$ and subsequently exported to the \texttt{JAX} ecosystem for seamless deployment. This unified framework enables batched network inference to be executed synchronously with the MPPI stochastic sampling loop, maximizing computational throughput. The MPPI hyperparameters employed in our experiments are detailed in Table~\ref{tab:mppi_param}.

\subsubsection{Gate-SDF Robustness and Spatial Consistency}
\begin{figure}[h]
    \centering
    \includegraphics[width=0.49\textwidth]{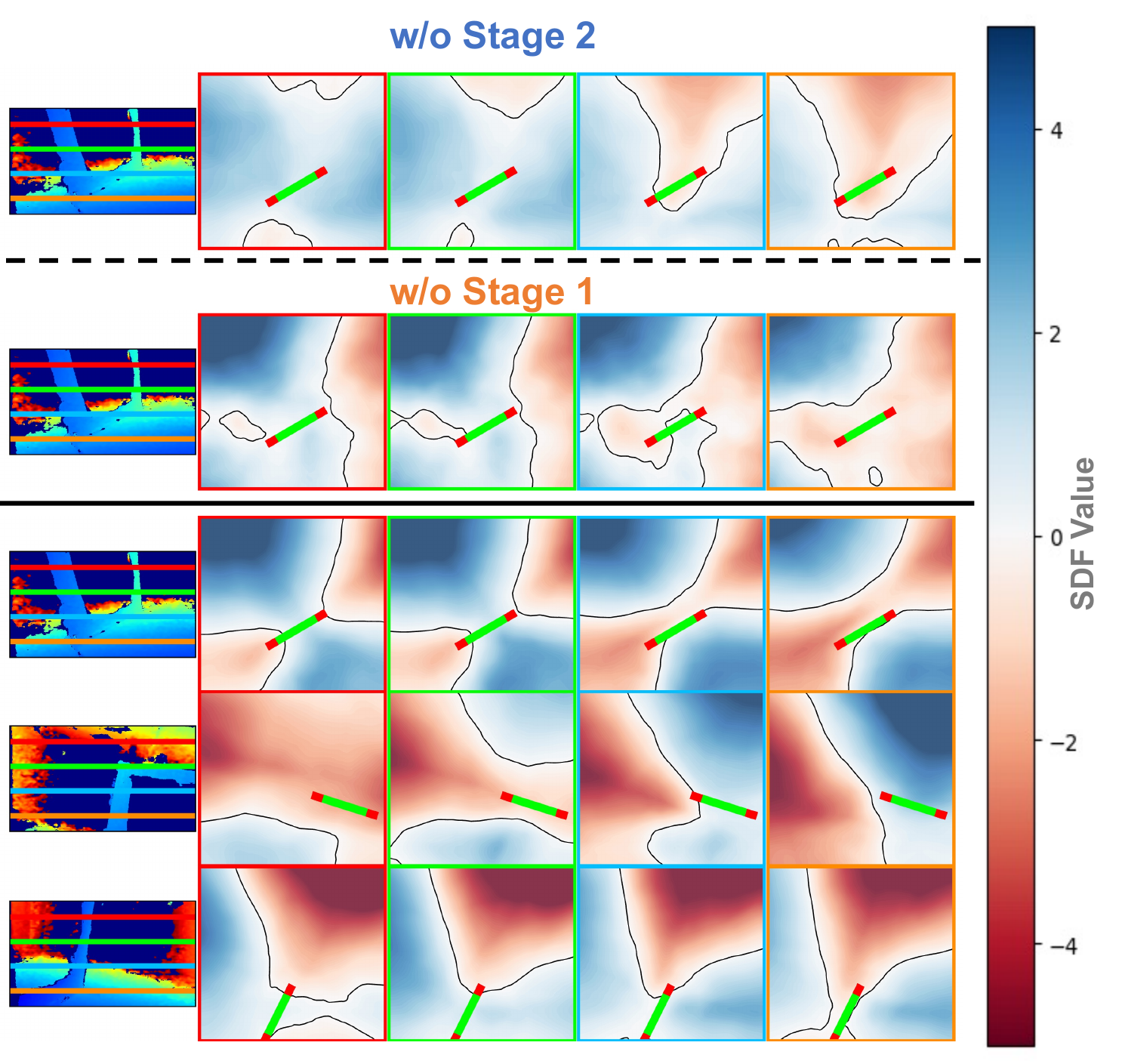}
    \caption{Predicted Gate-SDF visualization. The leftmost panel is the onboard depth image. Four horizontal slices (colored lines) are sampled, and their top-down SDF maps show that the model correctly captures the gate's traversable region.}
    \label{fig:occ}
    \vspace{-1em}
\end{figure}

Unlike conventional PnP methods requiring unobstructed views of multiple corners, our depth-conditioned Gate-SDF robustly handles noisy, partially observable depth images. As shown in Fig.\ref{fig:occ}, we select real depth images with diverse occlusion patterns, generate SDF evaluations on slices at multiple heights, and annotate the gate's ground-truth pose in the camera frame, where green denotes the traversable region and red denotes the gate frame. Furthermore, the ablation study in Fig.\ref{fig:occ} validates our two-stage training pipeline. The Stage 1 model struggles to generalize to real-world sensor noise, resulting in erroneous SDF estimations. Conversely, while the Stage 2 model captures the broad spatial trend, it fails to accurately reconstruct the fine-grained traversable structure of the gate. Our complete two-stage approach successfully infers the accurate traversal direction and precisely delineates non-traversable boundaries.

\begin{figure}[h]
    \centering
    \includegraphics[width=0.49\textwidth]{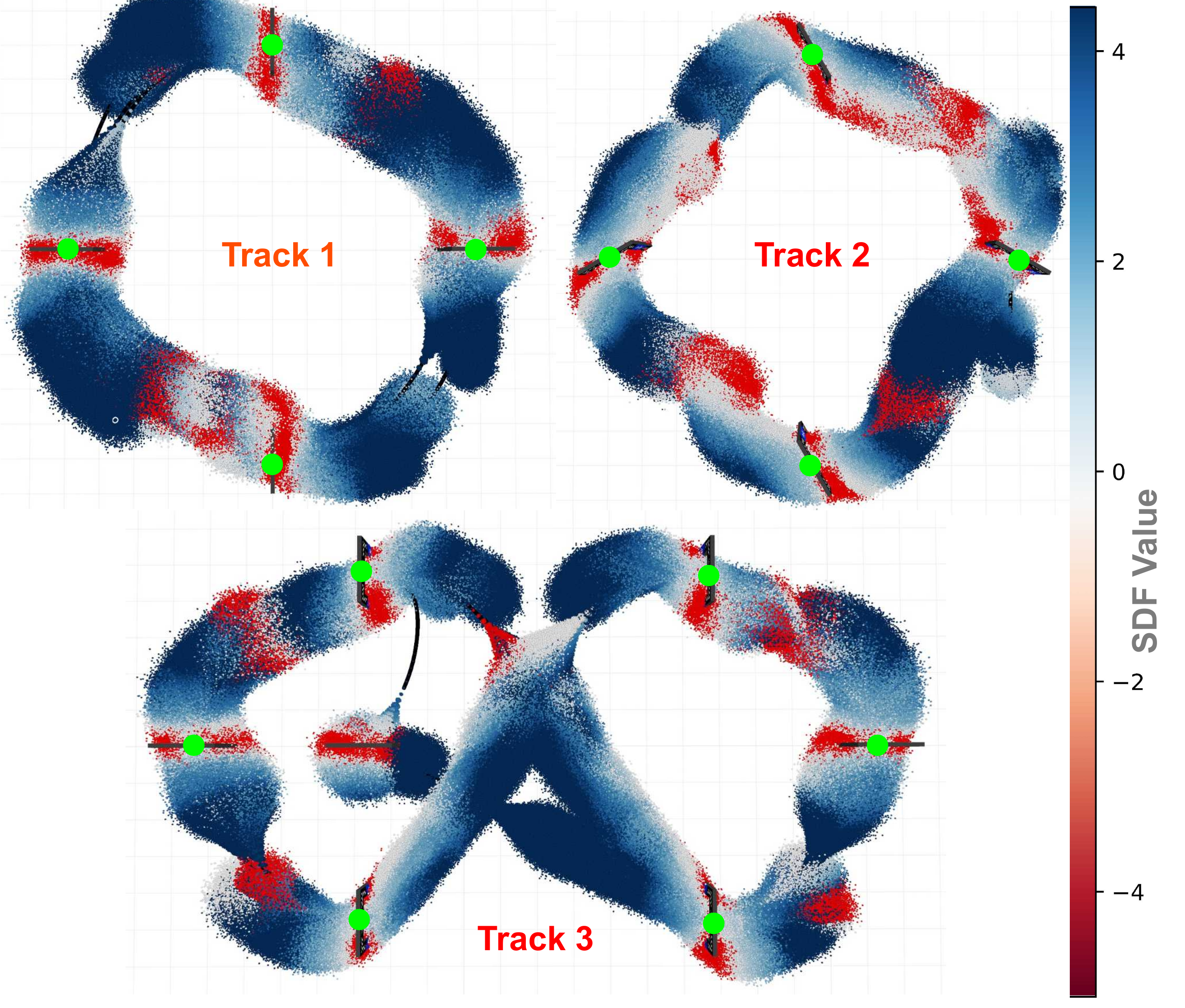}
    \caption{Visualization of the aggregated MPPI sampled trajectories over a complete flight lap.}
    \label{fig:consistency}
    \vspace{-1em}
\end{figure}

Furthermore, we evaluate the spatial consistency of the proposed representation throughout the dynamic flight process. We design three tracks, two of which share identical target waypoints but differ in gate orientation. As shown in Fig.\ref{fig:consistency}, we visualize the aggregated MPPI trajectory rollouts over \textbf{three} complete flight laps, color-coded by their predicted SDF values. The results demonstrate that the network maintains a highly consistent geometric representation of the environment over time. The predicted collision regions (red) accurately conform to the varying gate orientations, confirming that our method effectively generalizes to unseen spatial configurations and provides reliable, continuous guidance for the controller throughout the entire agile flight.

\begin{figure}[ht]
    \centering
    \includegraphics[width=0.49\textwidth]{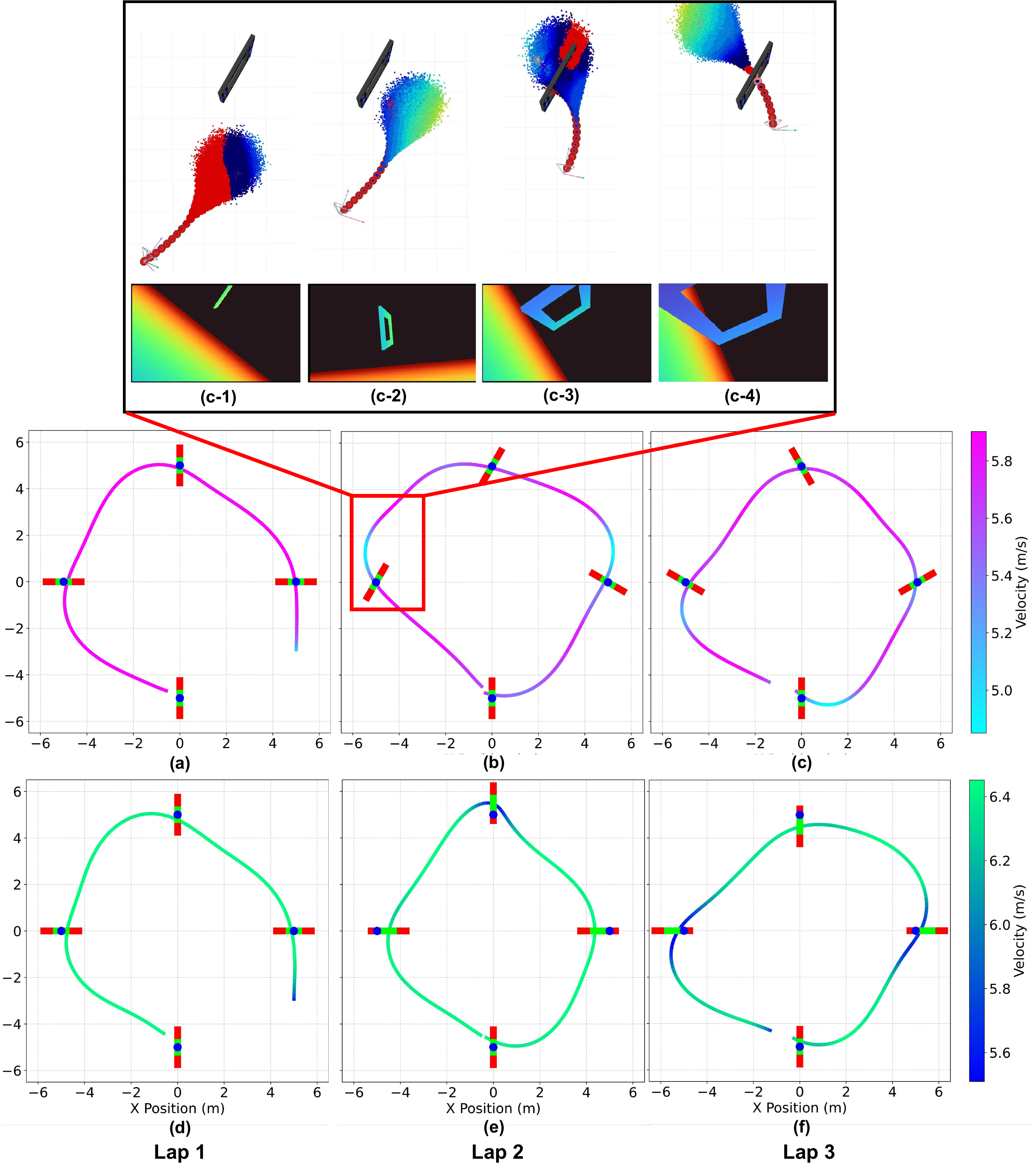}
    \caption{Visualization of continuous flight trajectories over three consecutive laps. Waypoints and gates are denoted by blue dots and solid lines, respectively. Subfigures (c-1) to (c-4) provide a detailed progression of the MPPI optimization process for the trajectory segment highlighted in red in (c).
     }
    \label{fig:trajectory}
    \vspace{-1.5em}
\end{figure}
\subsubsection{Simulation Scenarios}
To evaluate the performance of the proposed MPPI controller augmented with the learned Gate-SDF, simulation scenarios with varying difficulty levels are constructed. The method is assessed from two complementary perspectives on a circular track consisting of $4$ gates, as shown in Fig.\ref{fig:trajectory}, over three consecutive laps.

\begin{figure}[h]
    \centering
    \includegraphics[width=0.49\textwidth]{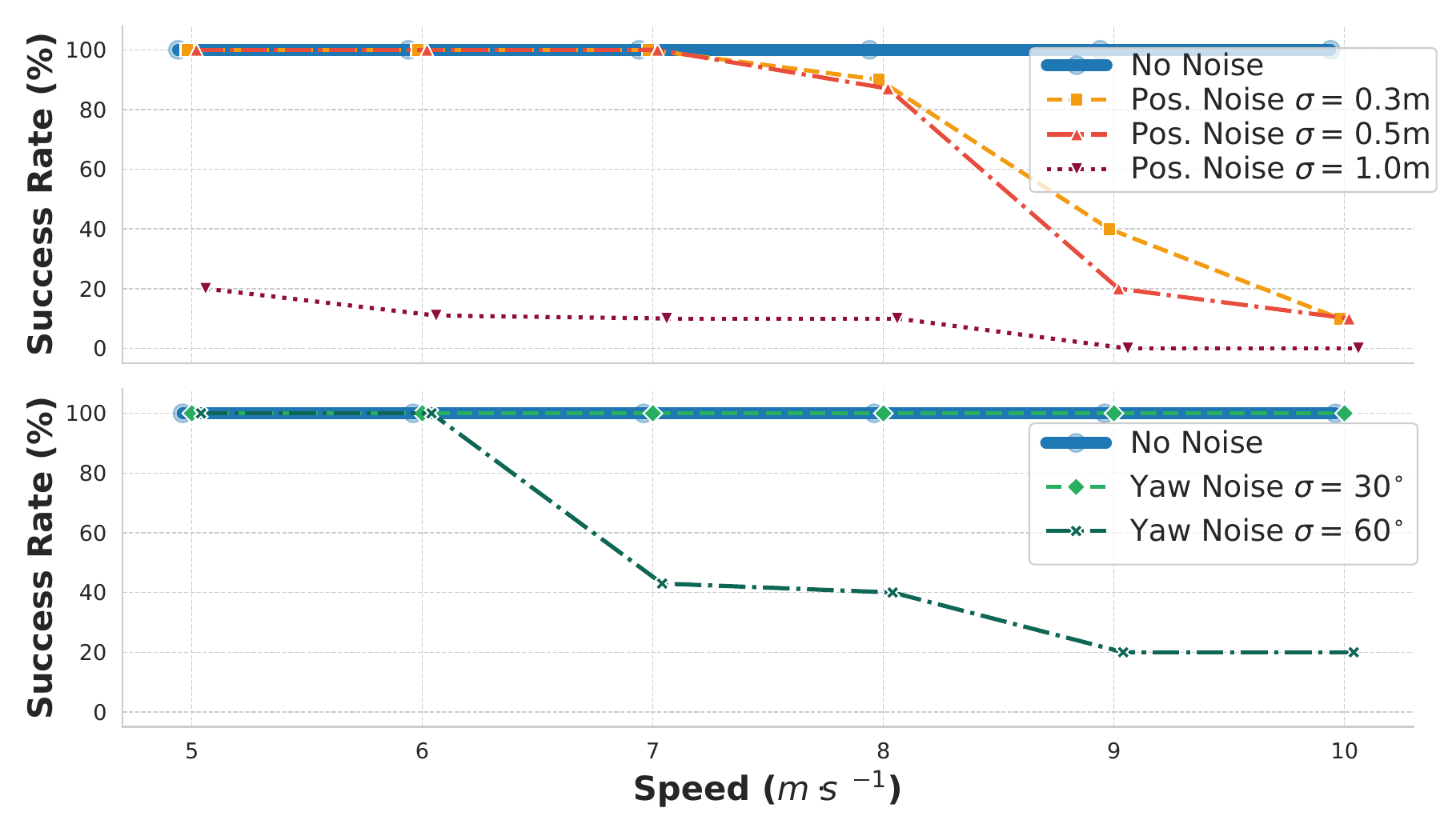}
    \caption{Success rates across different track configurations.
     }
    \label{fig:succ}
    \vspace{-1em}
\end{figure}

We evaluate the framework's robustness across two challenging conditions: \textbf{Scenario 1} applies random position perturbations to the true gates relative to the nominal waypoints, and \textbf{Scenario 2} applies random orientation perturbations, requiring anticipatory approach adjustments. Success rates across varying flight speeds and perturbation magnitudes are reported in Fig.\ref{fig:succ}. Lap trajectory visualizations (Fig.\ref{fig:trajectory}a--f) demonstrate that the controller successfully adapts to both disturbance types using solely onboard perception. 

\begin{table}[ht]
    \centering
    \caption{Performance Comparison with Baseline under Spatial Perturbations}
    \label{tab:comparison}
    \setlength{\tabcolsep}{3pt}
    \renewcommand{\arraystretch}{1.1}
    \scriptsize
    \resizebox{\columnwidth}{!}{%
    \begin{tabular}{l|cc|cc}
        \toprule
        \multirow{2}{*}{\textbf{Perturbation}} & \multicolumn{2}{c|}{\textbf{\cite{11297752} (Best Env.)}} & \multicolumn{2}{c}{\textbf{Ours}} \\
        & \textbf{SR} & \textbf{Max Vel} & \textbf{SR} & \textbf{Max Vel} \\
        \midrule
        No Noise & 100\% & 5.3 & \textbf{100\%} & \textbf{10.0} \\
        Pos. Noise ($\sim 0.3$\,m) & 100\% & 5.1 & \textbf{100\%} & \textbf{7.0}$^*$ \\
        Pos. Noise ($\sim 0.5$--$0.6$\,m) & 100\% & 4.6 & \textbf{100\%} & \textbf{7.0}$^\dagger$ \\
        Yaw Noise ($30^\circ$) & - & - & \textbf{100\%} & \textbf{10.0} \\
        Yaw Noise ($60^\circ$) & - & - & \textbf{100\%} & \textbf{6.0} \\
        \bottomrule
    \end{tabular}%
    }
    {\footnotesize $^*$ Ours maintains a $90\%$ SR at $8.0$\,m/s.}\\
    {\footnotesize $^\dagger$ Ours maintains an $87\%$ SR at $8.0$\,m/s.}
    \vspace{-2em}
\end{table}

As shown in Fig.\ref{fig:succ}, our method achieves a 100\% success rate across all speeds under nominal conditions. Under moderate position noise, Gate-SDF reactively adjusts the trajectory to maintain high success rates. However, excessive position perturbations (e.g., $1.0\,\text{m}$, exceeding the gate width) cause conflicts between the MPPI waypoint objective and visual observations, degrading performance. For orientation disturbances, the framework sustains 100\% success under minor noise, while larger angular perturbations reduce success rates, particularly at higher speeds where anticipatory maneuvering time is limited. Table \ref{tab:comparison} provides a quantitative reference against a recent vision-based RL baseline \cite{11297752}. Because the baseline operates in obstacle-cluttered environments, a direct comparison is inherently unfair; it serves solely to contextualize our framework's absolute speed and agility under spatial disturbances rather than claiming general superiority over RL methods.

To elucidate how the learned Gate-SDF governs trajectory optimization, Fig.~\ref{fig:trajectory}(c-1)--(c-4) visualizes the aggregated MPPI rollouts colored by their predicted SDF values, where red indicates physical collisions ($\hat{s} \le 0$). Crucially, during highly acute approaches, the lateral posts visually overlap, causing the gate to collapse into a single line in the depth image (e.g., Fig.~\ref{fig:trajectory}(c-1)). While traditional Perspective-n-Point (PnP) methods fail catastrophically here due to occluded corners, our Gate-SDF implicitly resolves this geometric ambiguity. It successfully infers the valid safety area directly from the collapsed depth profile, providing continuous spatial gradients that warp the MPPI sampling distribution to safely guide the vehicle through the narrow opening.

\begin{figure*}[t]
    \centering
    \includegraphics[width=0.99\textwidth, trim=0 2cm 0 1cm, clip]{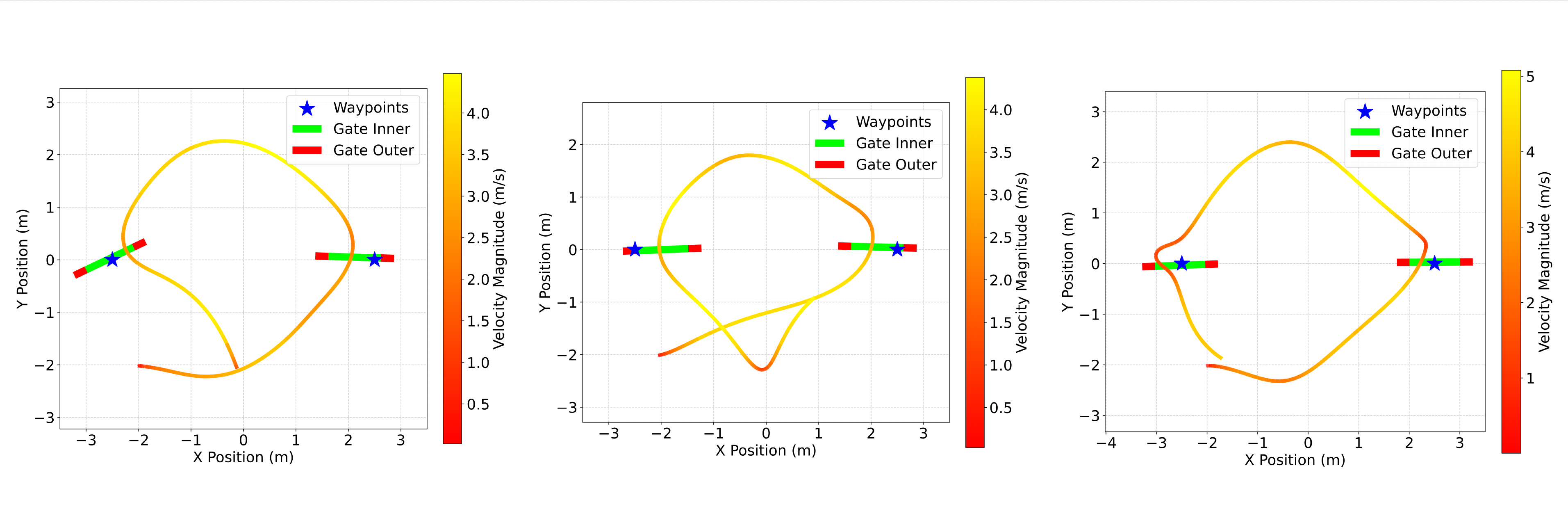}
    \caption{Trajectories from real-world experiments. The racing gates in each track configuration are subjected to position and orientation perturbations. Green denotes the gate traversable region, and red denotes collision regions.}
    \label{fig:real_trajectory}
    \vspace{-1em}
\end{figure*}
\subsection{Real‐world Experiments}
For the real-world validation, we employed a customized quadrotor with a total weight of $370\,\text{g}$ and a thrust-to-weight ratio (TWR) of $3$. All onboard computations were executed by an embedded Jetson Orin NX (16GB) module, utilizing a RealSense D435 depth camera for perception. State estimation for the quadrotor was provided by a motion-capture system, while the precise gate position and orientation were treated as unknown. Specifically, the depth image encoder operates in a separate PyTorch thread with a processing time of approximately $5\,\text{ms}$ using $\mathbf{z} \in \mathbb{R}^{64}$. The MPPI controller was configured with $M=1024$ rollouts and operated at a control frequency of approximately $50\,\text{Hz}$.

We first evaluated the dynamic performance of the proposed Gate-SDF-augmented MPPI controller on a circular racing track featuring two racing gates. Across multiple continuous flight trials, the gates' poses, encompassing both spatial positions and rotation angles, were randomly perturbed. The aggregated flight trajectories are depicted in Fig.\ref{fig:realtraj} and Fig.\ref{fig:real_trajectory}. Notably, the quadrotor achieved a maximum flight speed of $5.3\,\text{m/s}$ within this highly compact track, relying exclusively on onboard depth perception.

\begin{figure}[h]
    \vspace{-1em}
    \centering
    \includegraphics[width=0.49\textwidth]{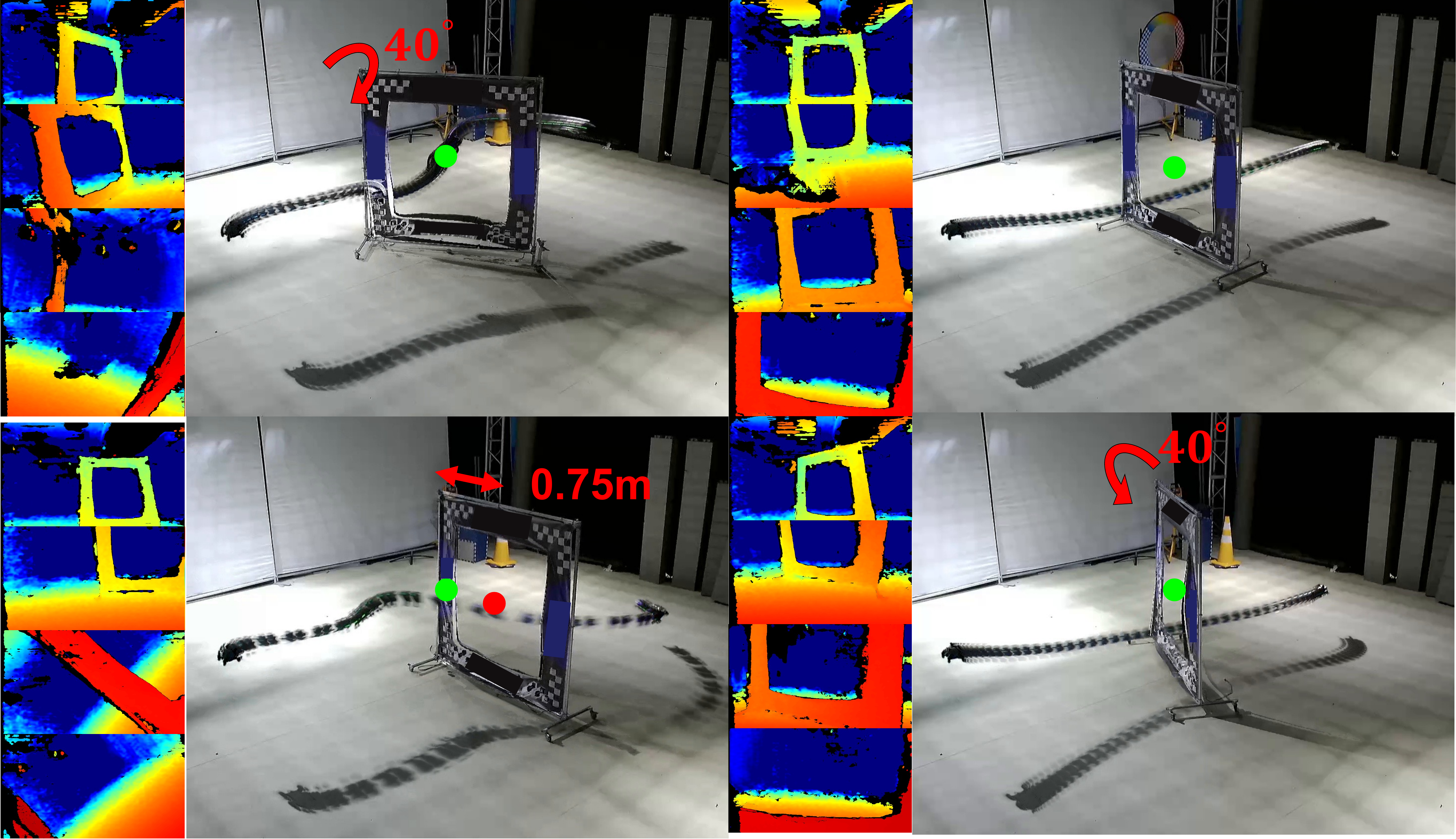}
    \caption{Flight trajectories of the point-to-point flight task.}
    \label{fig:real_singlegate}
    \vspace{-1em}
\end{figure}

To more explicitly demonstrate our method's ability to handle diverse gate placements and orientations, we subsequently designed a point-to-point flight task featuring a single gate deployed in various configurations. As illustrated in Fig.\ref{fig:real_singlegate}, the system successfully maneuvered through the racing gate even under severe initial displacement errors of up to $0.75\,\text{m}$ and orientation perturbations of up to $40^\circ$. These experimental results collectively highlight the robustness and efficacy of the proposed method in real-world environments subject to physical sensor noise and spatial uncertainties.

\section{Conclusion}
In this paper, we presented a novel vision-guided, reference trajectory free optimal control framework for autonomous drone racing. Instead of relying on engineered reference trajectories or explicit geometric priors, our approach extracts spatial constraints directly from onboard depth observations via a learned Neural Signed Distance Field (SDF). Integrating this representation with a Model Predictive Path Integral (MPPI) controller leverages GPU parallelism to evaluate complex safety boundaries across thousands of predictive rollouts in real time. Extensive validations demonstrate highly competitive agile flight and exceptional zero-shot robustness against uncalibrated gate positions, severe occlusions, and arbitrary orientations scenarios where traditional pose estimation often fails. Ultimately, this work bridges raw spatial perception and high-frequency stochastic optimal control, paving the way for highly adaptable autonomous flight in complex environments.
\bibliography{reference}
\bibliographystyle{ieeetr}
\end{document}